\documentclass[10pt,twocolumn,letterpaper]{article}

\usepackage[main]{cvpr}

\def\paperID{379}
\def\confName{3DV}
\def\confYear{2027}

\definecolor{cvprblue}{rgb}{0.21,0.49,0.74}
\usepackage[pagebackref,breaklinks,colorlinks,allcolors=cvprblue]{hyperref}

\title{AG-EgoPose: Spatially Anchored Residual Correction with Action Context for Monocular Egocentric 3D Pose Estimation}

\author{
Md Mushfiqur Azam\\
The University of Texas at San Antonio\\
{\tt\small mdmushfiqur.azam@utsa.edu}
\and
John Quarles\\
The University of Texas at San Antonio\\
{\tt\small john.quarles@utsa.edu}
\and
Kevin Desai\\
The University of Texas at San Antonio\\
{\tt\small kevin.desai@utsa.edu}
}

\usepackage[T1]{fontenc}
\usepackage{tikz}
\usetikzlibrary{calc}
\usetikzlibrary{positioning}
\usepackage{multirow}
\usepackage{wrapfig}
\usepackage{array}

\begin{document}

\maketitle

\begin{abstract}
Monocular egocentric 3D pose estimation is difficult because severe
foreshortening, self-occlusion, and a restricted field of view often remove the
image evidence needed to recover the camera wearer's body. Temporal context can
resolve these ambiguities, but unconstrained fusion may also corrupt joints that
are already localized reliably from the current frame. We present AG-EgoPose,
an action-guided framework that treats temporal information as a bounded-gate
residual correction to a spatial pose estimate. A frozen heatmap network
produces joint heatmaps and feature-pyramid features, from which a spatial
transformer estimates an independently supervised 3D pose. The heatmaps further
guide joint-local pooling
of frozen DINOv2 patch tokens, while DINOv2 frame tokens are processed by an
Ego4D-pretrained ActionFormer to obtain action-scale temporal context. Each
spatial joint token, augmented with its local appearance feature, attends to the
temporal sequence and predicts a zero-initialized 3D residual. A per-joint gate,
derived from heatmap location and confidence statistics, bounds the residual
weight.
AG-EgoPose improves over the strongest evaluated baselines by 10.1\% PA-MPJPE
on EgoPW, and by 9.0\% MPJPE and 6.3\% PA-MPJPE on SceneEgo after
fine-tuning. Code will be made publicly available at \href{https://github.com/}{https://github.com/}.
\end{abstract}

% Introduction
\section{Introduction}
\label{sec:intro}

Egocentric 3D human pose estimation aims to recover the camera wearer's body from a
head-mounted camera and supports applications in AR/VR, avatar animation, and
human--computer interaction. A monocular headset is lightweight and requires no
external cameras, stereo rig, or body-worn markers. This convenience, however,
makes pose recovery severely underconstrained. The fisheye camera observes the
body from above, producing strong perspective distortion and foreshortening,
while self-occlusion and the limited field of view frequently hide the lower
body and arms. Consequently, methods developed for third-person images
~\cite{openpose,densepose,deep} transfer poorly to this setting.

Existing egocentric methods improve spatial representations
~\cite{xregopose,selfpose,egoattention}, incorporate scene geometry
~\cite{sceneego,visionspan}, exploit stereo views~\cite{unrealego,egoposeformer},
or model short pose sequences~\cite{egostan,egoformer}. Nevertheless, two
problems remain for monocular video. First, a heatmap may locate a visible joint
accurately but offers little evidence for an occluded or out-of-view joint.
Second, temporal fusion is not uniformly helpful: motion can disambiguate a
hidden leg from the gait cycle, yet a temporal branch with too much freedom can
degrade an already accurate spatial prediction. Temporal modeling should
therefore complement frame-level localization rather than replace it.

Our central observation is that these streams should have asymmetric roles.
Heatmap evidence provides a strong, directly supervised spatial anchor, whereas
action-level temporal context is better suited to correcting ambiguities in
that estimate. This motivates a residual formulation in which temporal features
predict a signed pose correction whose influence is modulated by heatmap
statistics from the spatial observation.

We introduce \emph{AG-EgoPose}, an action-guided framework for monocular
egocentric 3D pose estimation. A frozen ConvNeXt-Tiny~\cite{convnext} heatmap
network predicts 2D joint heatmaps and a dense FPN feature map. For each joint,
a tokenizer combines the heatmap with the FPN feature sampled at its soft
location, and a Spatial Joint Transformer models inter-joint dependencies and
predicts an independently supervised spatial pose $\mathbf{P}^{\mathrm{sp}}$. In
parallel, a frozen DINOv2~\cite{dinov2} encoder supplies patch and frame tokens:
the predicted heatmaps pool the patch tokens into joint-local appearance
descriptors, while the frame tokens feed an Ego4D-pretrained
ActionFormer~\cite{actionformer,ego4d} for action-scale temporal context.

Each spatial joint token, augmented with its local descriptor, attends over this
context and predicts a residual $\Delta\mathbf{P}$ whose output layer is
initialized to zero, so optimization begins exactly from the spatial estimator.
Heatmap moments, dispersion, entropy, and peak statistics separately determine a
bounded per-joint gate $\alpha$, giving
$\mathbf{P}=\mathbf{P}^{\mathrm{sp}}+\alpha\Delta\mathbf{P}$. Because the gate
sees no motion features, it stays tied to spatial evidence rather than becoming
an unconstrained routing variable. We evaluate on EgoPW~\cite{egopw} and
SceneEgo~\cite{sceneego}. Our contributions are:

\begin{itemize}
    \item We formulate temporal modeling as a bounded-gate residual correction to
    an independently supervised spatial pose, preserving frame-level evidence
    while using action context as a complementary correction signal.

    \item We introduce heatmap-guided joint representations at two complementary
    levels: FPN-aware spatial tokens for pose anchoring and joint-local DINOv2
    appearance tokens for querying action-scale temporal context.

    \item We design a zero-initialized correction decoder with a per-joint gate
    derived from heatmap location and confidence statistics, and validate the
    formulation on EgoPW and SceneEgo through targeted ablations.
\end{itemize}

\noindent AG-EgoPose reaches 63.0~mm PA-MPJPE on EgoPW and 90.8/68.8~mm
MPJPE/PA-MPJPE on SceneEgo, improving over the strongest evaluated baselines by
10.1\% and 9.0\%/6.3\%.

% Related Work
\section{Related Work}
\label{sec:related_work}

Egocentric 3D human pose estimation is challenging because of fisheye
distortion, self-occlusion, and limited body visibility
~\cite{egocentricsurvey}. Early systems such as EgoCap~\cite{egocap}
demonstrated the feasibility of head-mounted capture, and subsequent work has
explored monocular, stereo, scene-assisted, and temporal formulations.

\noindent\textbf{Monocular heatmap-based methods.}
Mo$^2$Cap$^2$~\cite{mo2cap2} established a real-time pipeline that predicts
2D joint heatmaps from a fisheye image and lifts them to 3D. EgoGlass
~\cite{egoglass} augments heatmap supervision with body-part segmentation and
pseudo-limb masks, while xR-EgoPose~\cite{xregopose} and SelfPose
~\cite{selfpose} improve lifting through dual-branch encoder--decoders and
probabilistic intermediate representations. EgoTAP~\cite{egoattention}
constructs joint descriptors from image and heatmap features and propagates
them with attention. Hong et al.~\cite{jswap} further introduce joint-specific
weighted pooling of heatmap evidence. These methods establish heatmaps as a
strong spatial representation, but most operate frame by frame and cannot use
action-scale context when joints are not observed.

\noindent\textbf{Scene-aware and weakly supervised methods.}
SceneEgo~\cite{sceneego} reconstructs scene depth and uses the recovered
geometry to constrain body pose, while Jiang and Grauman~\cite{invisiblepose}
combine scene structure and motion to recover joints that are never directly
observed. Jiang and Ithapu~\cite{visionspan} instead estimate pose from the
scene content within the wearer's vision span, and Wang et al.
~\cite{egoglobal} recover egocentric pose in a global scene coordinate frame.
EgoPW~\cite{egopw} instead uses a synchronized third-person camera to obtain
weak 3D supervision for in-the-wild monocular video. These methods improve
ambiguity resolution through additional geometry or supervision, whereas our
inference pipeline requires only monocular RGB video.

\noindent\textbf{Multi-view and visibility-aware transformers.}
Stereo systems reduce depth ambiguity through paired fisheye views. UnrealEgo
~\cite{unrealego} introduced a large synthetic stereo benchmark and a
dual-branch heatmap-and-pose network, and Ego3DPose~\cite{ego3dpose} combines
stereo matching with perspective-aware heatmaps. EgoPoseFormer
~\cite{egoposeformer} refines coarse joint proposals using deformable
cross-attention over stereo image features. Its successor, EgoPoseFormer v2
~\cite{egoposeformerv2}, adds identity-conditioned queries, multi-view spatial
refinement, causal temporal attention, and uncertainty-aware semi-supervised
training for temporally consistent headset-based motion estimation. EvaPose
~\cite{evapose} addresses invisibility explicitly by learning from the keypoint
visibility annotations introduced with the Eva-3M dataset. These approaches
highlight the importance of temporal consistency and uncertainty, but target
multi-view capture or rely on explicit visibility supervision. Our method
instead derives a correction gate directly from predicted monocular heatmap
statistics.

\noindent\textbf{Temporal egocentric pose estimation.}
EgoSTAN
~\cite{egostan} and EgoFormer~\cite{egoformer} apply spatiotemporal attention
to image or joint tokens, and EgoCast~\cite{egocast} extends temporal modeling
to future-pose prediction. Other methods use external context: Ego+X
~\cite{ego+x} models social interactions, while You2Me~\cite{you2me} infers
the wearer's pose from a visible interacting person. In contrast, AG-EgoPose
repurposes an ActionFormer~\cite{actionformer} pretrained for temporal action
localization on Ego4D~\cite{ego4d}. Rather than directly regressing pose from
the temporal representation, it uses this representation only to predict a
bounded correction to a supervised spatial estimate.

\noindent\textbf{Mesh and motion reconstruction.}
Fisheye-aware methods also estimate richer body representations. Fish2Mesh
~\cite{fish2mesh}, EgoWholeBody~\cite{egowholebody}, and EgoHMR
~\cite{egohmr} recover parametric body meshes using fisheye transformers or
diffusion refinement. EgoEgo~\cite{egoego} conditions body motion on head
trajectory, EgoAllo~\cite{egoallo} estimates body and hand motion from
egocentric SLAM, and REWIND~\cite{rewind} generates real-time whole-body motion
with exemplar-based identity conditioning. These methods address mesh or motion
reconstruction, whereas we focus on monocular per-frame 3D joint estimation
with video context.

\begin{figure*}[t]
\centering
\includegraphics[width=\textwidth]{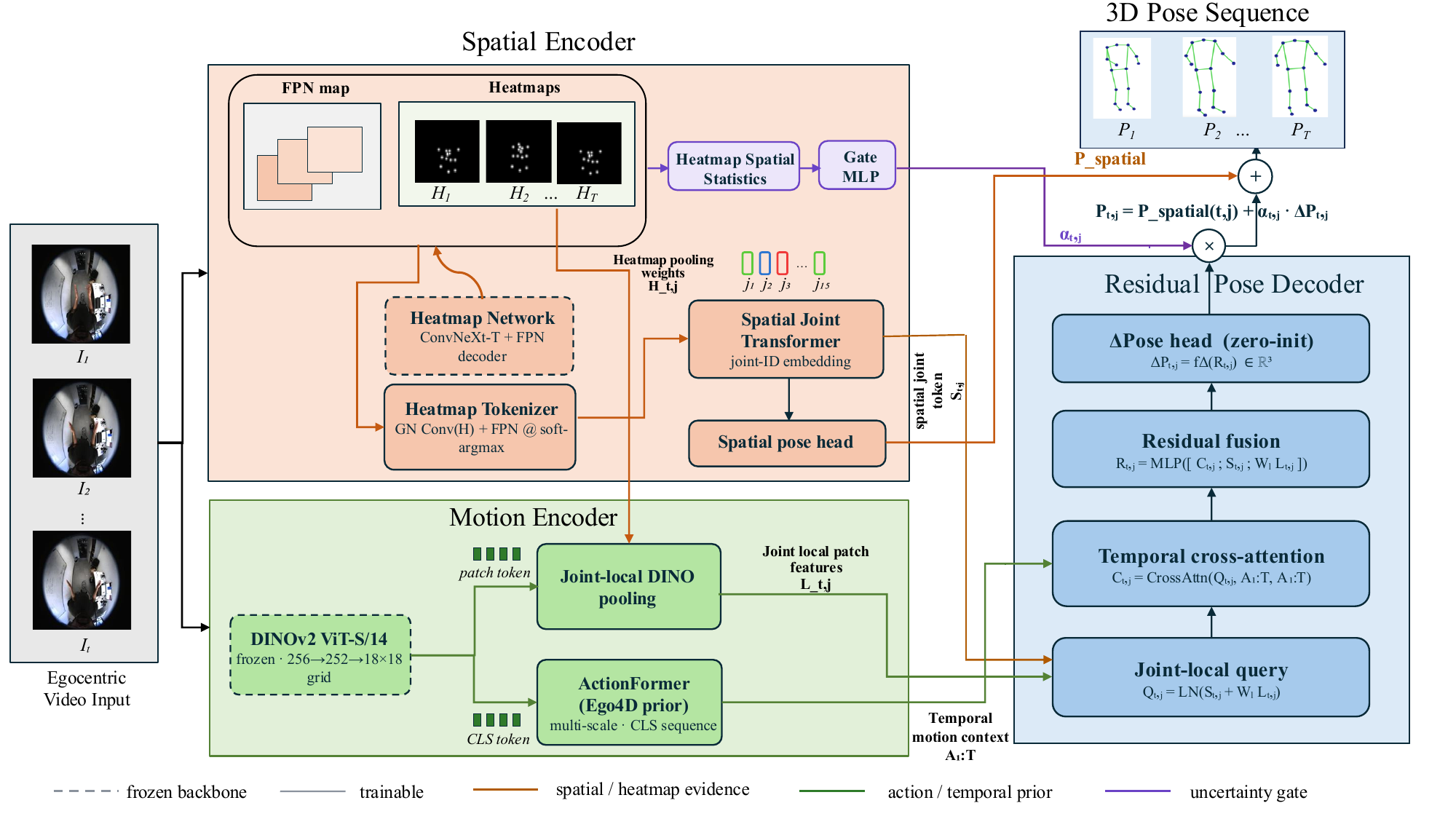}
\vspace{-7mm}
\caption{Overview of AG-EgoPose. A frozen heatmap network produces joint
heatmaps and FPN features. Their joint tokens are refined by a Spatial Joint
Transformer and mapped to the spatial pose $\mathbf{P}^{\mathrm{sp}}$.
Predicted heatmaps also pool frozen DINOv2 patch tokens into joint-local
appearance features, while DINOv2 frame tokens are encoded by an
Ego4D-pretrained ActionFormer. Joint-local queries attend to this temporal
context to predict a zero-initialized residual $\Delta\mathbf{P}$. Heatmap
location and confidence statistics independently produce a bounded gate
$\alpha$, yielding
$\mathbf{P}=\mathbf{P}^{\mathrm{sp}}+\alpha\Delta\mathbf{P}$.}
\label{fig:architecture}
\end{figure*}

% Method
\section{Method}
\label{sec:method}

Given an egocentric RGB clip $\mathbf{I}=\{I_t\}_{t=1}^{T}$, AG-EgoPose
predicts camera-frame 3D body joints
$\mathbf{P}\in\mathbb{R}^{T\times J\times3}$ with $J{=}15$. The model is built
around a simple design choice: a pose estimated from the current image should
remain the anchor, and temporal evidence should only be allowed to correct it.
This avoids forcing an action representation to explain image-specific pose
details that are better recovered from heatmaps. As shown in
Fig.~\ref{fig:architecture}, we first estimate an independently supervised
spatial pose $\mathbf{P}^{\mathrm{sp}}$, then predict a gate-weighted residual:
\begin{equation}
    \mathbf{P}_{t,j}
    = \mathbf{P}^{\mathrm{sp}}_{t,j}
    + \alpha_{t,j}\Delta\mathbf{P}_{t,j}.
    \label{eq:pose_correction}
\end{equation}
Here $\Delta\mathbf{P}_{t,j}$ is a signed 3D correction and
$\alpha_{t,j}$ controls the residual contribution for each joint and frame. We
bound the gate rather than the residual vector itself; the
residual magnitude is controlled by zero initialization and a soft residual
regularizer.
\subsection{Spatial Pose Anchor}
\label{subsec:spatial_encoder}

\subsubsection{Heatmap Evidence}
\label{subsubsec:heatmap_evidence}

\begin{figure}[t]
\centering
\includegraphics[width=\linewidth]{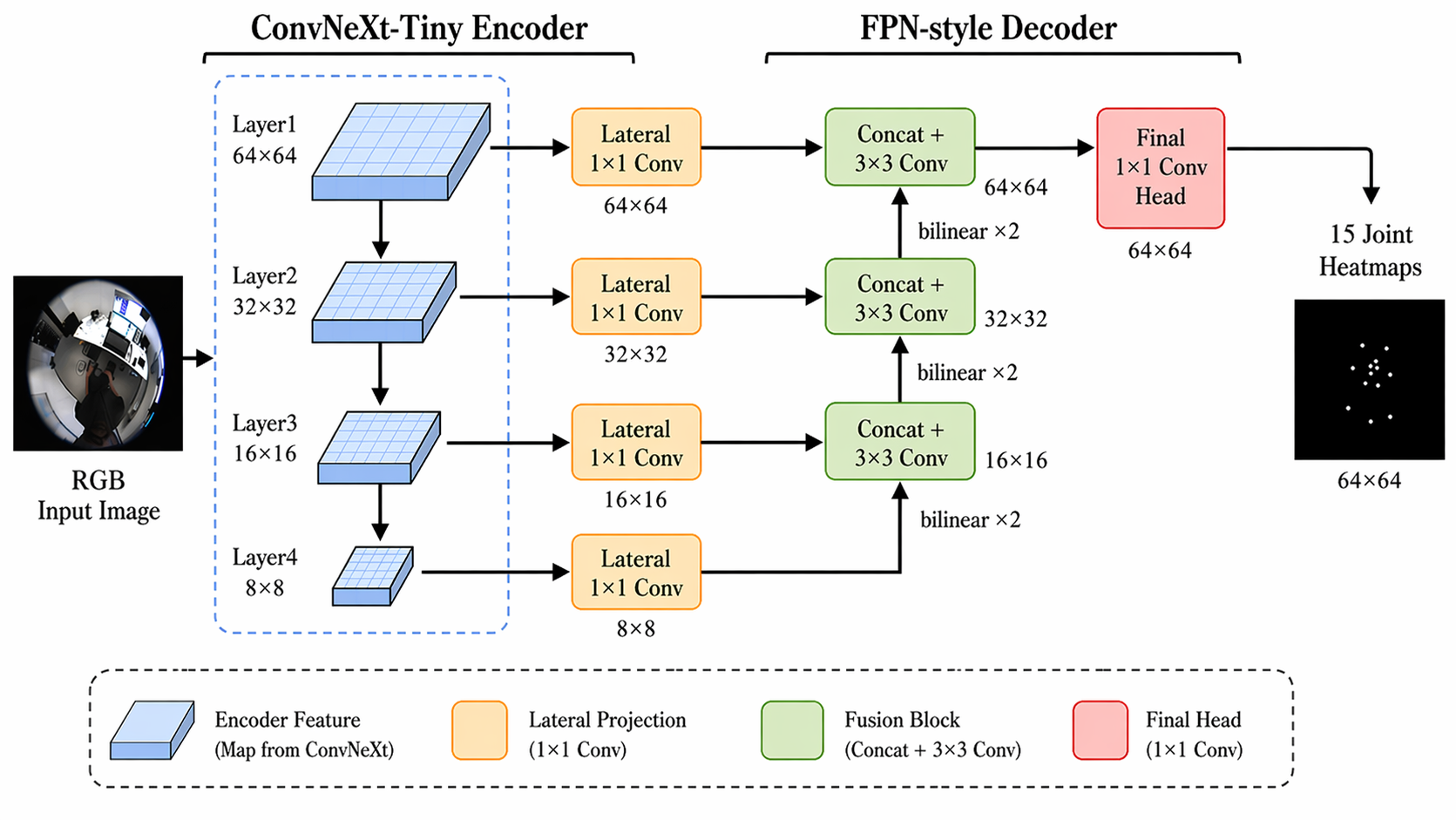}
\caption{Frozen 2D heatmap network. A ConvNeXt-Tiny encoder and FPN-style
decoder produce $J$ heatmap logits and a $96$-channel feature map at
$64{\times}64$ resolution. Both outputs are reused by the 3D pose model.}
\label{fig:heatmap}
\end{figure}

As shown in Fig.~\ref{fig:heatmap}, each frame is resized to
$256{\times}256$ and processed by a ConvNeXt-Tiny encoder with an FPN-style
decoder~\cite{convnext,fpn}. The heatmap network outputs $J$ heatmap logits and
a $96$-channel FPN feature map at
$64{\times}64$ resolution. We train this network with 2D joint supervision and
freeze it during 3D pose training, giving the pose model a stable spatial
coordinate system for 3D reconstruction.

\subsubsection{Heatmap Tokenization}
\label{subsubsec:heatmap_tokenization}

For each joint, we encode the predicted heatmap with a small convolutional
tokenizer using GroupNorm and GELU. The heatmap itself captures the spatial
belief over the joint, including multi-modal or diffuse responses. To attach
image evidence at the same location, we compute a soft heatmap centroid and
bilinearly sample the frozen FPN feature map at that point. The heatmap token
and sampled FPN feature are projected to the same dimension and fused with
LayerNorm, producing a joint token
$\mathbf{s}^{0}_{t,j}\in\mathbb{R}^{128}$. The token therefore carries both the
predicted joint distribution and local appearance from the heatmap network.

\subsubsection{Joint Reasoning and Spatial Pose}
\label{subsubsec:spatial_joint_reasoning}

The per-joint tokens are still local at this stage. We therefore add a learned
joint identity embedding and pass the $J$ tokens of each frame through a
one-layer Spatial Joint Transformer. This self-attention layer lets the model
compare joints within the same frame, which helps with left-right identity and
kinematic consistency before any temporal information is used. A lightweight MLP
head maps the refined token $\mathbf{s}_{t,j}$ to
$\mathbf{P}^{\mathrm{sp}}_{t,j}$. We supervise this spatial output directly, so
the model maintains an image-based pose estimate even when temporal correction
is weak.

\subsection{Joint-Local Appearance and Action Context}
\label{subsec:motion_encoder}

The correction branch uses two complementary signals from a frozen DINOv2
ViT-S/14~\cite{dinov2}. Patch tokens provide local appearance around each joint,
while the CLS token sequence provides a compact frame-level description for
temporal modeling. We resize frames to $252{\times}252$, giving an
$18{\times}18$ patch grid and $384$-D DINOv2 tokens.

\subsubsection{Heatmap-Guided DINO Pooling}
\label{subsubsec:heatmap_guided_dino_pooling}

A global DINO token is not enough to tell the decoder whether the correction is
about a wrist, knee, or ankle. We therefore use the predicted heatmap as a
soft spatial query over the DINOv2 patch grid. For joint $j$, the heatmap is
resized to the patch resolution, normalized with a softmax, and mixed with a
small uniform prior:
\begin{equation}
    \mathbf{l}_{t,j}
    = \sum_{n=1}^{N}
      \left((1-\rho)\,\mathrm{softmax}(\tau\bar{\mathbf{H}}_{t,j})_n
      + \frac{\rho}{N}\right)\mathbf{D}_{t,n},
    \label{eq:dino_pooling}
\end{equation}
where $\mathbf{D}_{t,n}$ is the $n$-th DINOv2 patch token. We use
$\tau{=}6$ and $\rho{=}0.15$. The uniform mixture is a small safeguard against
over-trusting a displaced heatmap; it preserves some context while keeping the
feature joint-local.

\subsubsection{ActionFormer Temporal Context}
\label{subsubsec:actionformer_temporal_context}

For temporal context, we feed the normalized DINOv2 CLS sequence to an
ActionFormer~\cite{actionformer} pretrained for temporal action localization on
Ego4D~\cite{ego4d}. Unlike a frame-wise temporal MLP, this backbone was trained
to identify actions with variable duration. Its features therefore encode both
local transitions and longer activity phases, which are useful when the pose of
an occluded joint cannot be determined from the current frame alone.

The pretrained ActionFormer \cite{actionformer} expects $256$-D EgoVLP \cite{egovlp} descriptors, whereas our
DINOv2 CLS tokens are $384$-D. We bridge the two representations with a
learned temporal $1{\times}1$ projection and pass the sequence through the
pretrained stem and branch blocks:
\begin{equation}
\begin{aligned}
    \mathbf{V}
    &= [\mathrm{LN}(\mathbf{d}^{\mathrm{cls}}_1),\ldots,
        \mathrm{LN}(\mathbf{d}^{\mathrm{cls}}_T)]^{\top}, \\
    \mathbf{X} &= f_{\mathrm{in}}(\mathbf{V}^{\top}), \\
    \mathbf{B}^{(0)} &= \Phi_{\mathrm{stem}}(\mathbf{X}), \\
    \mathbf{B}^{(\ell)} &= \Phi_{\ell}(\mathbf{B}^{(\ell-1)}),
        \quad \ell=1,\ldots,7.
\end{aligned}
\label{eq:actionformer_pyramid}
\end{equation}
Here $\mathbf{X}\in\mathbb{R}^{256\times T}$ and
$\mathbf{B}^{(\ell)}\in\mathbb{R}^{384\times T_{\ell}}$. ActionFormer builds a
temporal pyramid with masked 1D embedding convolutions and multi-head
convolutional attention. Early blocks operate at higher temporal resolution,
while later blocks aggregate longer-range context, giving the residual decoder
both short-term motion cues and coarser action phase information.

The original localization heads classify actions and regress their temporal
boundaries; we discard them. Since pose estimation needs context at every input
frame, we resize all backbone outputs to length $T$ and fuse them along the
channel dimension:
\begin{equation}
\begin{aligned}
    \mathbf{C} &=
      \mathop{\Vert}_{\ell=0}^{7}
      \mathrm{Up}_{T}(\mathbf{B}^{(\ell)}), \\
    \mathbf{A} &=
      [f_{\mathrm{fuse}}(\mathbf{C})]^{\top}
      \in\mathbb{R}^{T\times384}.
\end{aligned}
    \label{eq:actionformer_fusion}
\end{equation}
where $\Vert$ denotes channel-wise concatenation and
$\mathrm{Up}_{T}$ is nearest-neighbor interpolation to length $T$.
$f_{\mathrm{fuse}}$ is a temporal $1{\times}1$ projection. Thus,
$\mathbf{A}_t$ combines fine and coarse action evidence but remains aligned
with frame $t$ for joint-specific retrieval. DINOv2 remains frozen. During pose
training we adapt only the projection layers and the last two ActionFormer
blocks; the remaining pretrained blocks stay frozen, so the transferred Ego4D
representation is preserved rather than retrained
(Sec.~\ref{sec:implementation}).

\subsection{Heatmap-Statistics-Gated Residual Decoder}
\label{subsec:pose_decoder}

\subsubsection{Per-Joint Temporal Retrieval}
\label{subsubsec:per_joint_temporal_retrieval}

The residual decoder asks a different temporal question for each joint. We first
project the joint-local DINO feature $\mathbf{l}_{t,j}$ to the spatial token
dimension and add it to the refined heatmap/FPN token:
\begin{equation}
    \mathbf{q}_{t,j}
    = \mathrm{LN}\!\left(\mathbf{s}_{t,j}+f_l(\mathbf{l}_{t,j})\right).
    \label{eq:local_query}
\end{equation}
For a fixed joint $j$, the sequence $\{\mathbf{q}_{t,j}\}_{t=1}^{T}$ attends to
the ActionFormer context sequence $\{\mathbf{A}_t\}_{t=1}^{T}$. This lets a knee
query, for example, retrieve different temporal evidence from a wrist query,
even though both share the same action-context memory. The attended feature is
concatenated with the spatial token and projected local appearance, then passed
through an MLP to predict $\Delta\mathbf{P}_{t,j}$. The last linear layer of the
residual head is initialized to zero. Thus, at initialization the model is
exactly the spatial anchor in Eq.~\eqref{eq:pose_correction}, and the correction
is learned only when it reduces the pose loss.

\subsubsection{Heatmap-Statistics Gate}
\label{subsubsec:heatmap_statistics_gate}

The residual contribution should be conditioned on the spatial evidence that
anchors the pose. We therefore compute the gate from heatmap statistics: soft
centroid, variance in $x$ and $y$,
covariance, entropy, peak response, and the gap between the two largest
responses. These statistics summarize both location and confidence without
introducing another image encoder. A small MLP maps them to a bounded gate,
\begin{equation}
\begin{aligned}
    \alpha_{t,j}
    &= \alpha_{\min}
      +(\alpha_{\max}-\alpha_{\min})
       \sigma(f_{\alpha}(\mathbf{u}_{t,j})), \\
    \alpha_{t,j} &\in[0.05,0.80].
\end{aligned}
\label{eq:alpha_gate}
\end{equation}
This gate range is deliberately conservative. It limits the multiplicative
weight assigned to the temporal residual, while the residual norm is discouraged
by the training objective. Together, these choices reduce the risk that the
action branch overwrites the spatial pose during dataset adaptation. The gate
does not directly estimate temporal reliability; it modulates correction from the
heatmap-derived spatial state.

\subsection{Training Objective}
\label{subsec:loss}

We supervise both the final prediction and the spatial anchor. The final MPJPE
optimizes the output used at test time, while the spatial MPJPE keeps
$\mathbf{P}^{\mathrm{sp}}$ a reliable fallback for the residual formulation.
Coordinate loss alone, however, can still produce limbs with plausible joint
positions but inconsistent local geometry. We therefore add two lightweight
kinematic terms over the skeleton edge set $\mathcal{E}$:
\begin{equation}
\begin{aligned}
\mathcal{L}_{\mathrm{bone}}
&= \frac{1}{|\mathcal{E}|}
   \sum_{(p,j)\in\mathcal{E}}
   \left(\|\mathbf{b}_{p,j}\|_2-\|\mathbf{b}^{*}_{p,j}\|_2\right)^2,\\
\mathcal{L}_{\mathrm{dir}}
&= 1-\frac{1}{|\mathcal{E}|}
   \sum_{(p,j)\in\mathcal{E}}
   \frac{\mathbf{b}_{p,j}^{\top}\mathbf{b}^{*}_{p,j}}
        {\|\mathbf{b}_{p,j}\|_2\|\mathbf{b}^{*}_{p,j}\|_2}.
\end{aligned}
\label{eq:bone_direction_loss}
\end{equation}
where $\mathbf{b}_{p,j}=\mathbf{P}_{j}-\mathbf{P}_{p}$ and $*$ denotes ground
truth. The bone-length term preserves limb scale, and the direction term
encourages each limb to point toward the correct 3D orientation.

The complete objective is
\begin{equation}
\begin{aligned}
    \mathcal{L}
    &= \mathcal{L}_{\mathrm{final}}
     + \lambda_{\mathrm{sp}}\mathcal{L}_{\mathrm{sp}}
     + \lambda_{\mathrm{bone}}\mathcal{L}_{\mathrm{bone}} \\
    &\quad
     + \lambda_{\mathrm{dir}}\mathcal{L}_{\mathrm{dir}}
     + \lambda_{\mathrm{res}}\mathcal{L}_{\mathrm{res}}
     + \lambda_{\alpha}\mathcal{L}_{\alpha}.
\end{aligned}
\label{eq:training_loss}
\end{equation}
Here $\mathcal{L}_{\mathrm{res}}$ is the mean norm of $\Delta\mathbf{P}$ and
$\mathcal{L}_{\alpha}$ is the mean correction gate; both discourage corrections
larger than the spatial evidence warrants. Loss weights are listed in
Sec.~\ref{sec:implementation}. The 2D heatmap network and DINOv2 backbone remain
frozen during 3D training.

% Implementation Details
\section{Experimental Setup}
\label{sec:implementation}
\subsection{Datasets}
\label{subsec:training_dataset}
We evaluate on EgoPW~\cite{egopw} and SceneEgo~\cite{sceneego}. EgoPW is our
primary training benchmark: it contains over 318K in-the-wild egocentric frames
from 10 subjects performing 20 everyday activities, with pseudo-3D labels
obtained by fusing head-mounted fisheye video with synchronized external-view
observations. SceneEgo contains approximately 28K frames from two subjects and
provides accurate 3D joint annotations from a calibrated multi-view motion
capture setup. We use SceneEgo for pretrain-then-finetune transfer by
adapting the EgoPW-pretrained model on the official SceneEgo training split.

\subsection{Implementation Details}

\noindent\textbf{2D heatmap training.}
We train the ConvNeXt-Tiny heatmap network on EgoPW and projected SceneEgo
training frames to predict $J{=}15$ joint heatmaps from $256{\times}256$
frames. The heatmap
objective combines a CenterNet-style Gaussian focal loss, BCE-with-logits, and
a soft-argmax coordinate loss, with weights $1.0$, $0.10$, and $5.0$,
respectively. Target heatmaps are Gaussians with $\sigma{=}2$ centered at the
2D joint locations, and the coordinate term compares soft-argmax locations from
predicted and target maps. SceneEgo does not provide 2D labels, so we project
its 3D joints using the provided fisheye calibration. We train for 30 epochs
with batch size 32 using AdamW (weight decay $10^{-4}$), learning rates of
$10^{-5}$ for the pretrained backbone and $10^{-4}$ for the decoder and heatmap
head, two warm-up epochs followed by cosine decay, and photometric augmentation
(color jitter, occasional blur and grayscale, and random erasing). The heatmap
network is frozen for all 3D experiments.
\begin{table*}[t]
\centering

\begin{minipage}[t]{0.48\textwidth}
\centering
\small
\begin{tabular}{l|c}
\hline
\textbf{Method} & \textbf{PA-MPJPE}$\downarrow$ \\
\hline
SelfPose~\cite{selfpose} & 112.0 \\
Kolotouros et al.~\cite{probabilistichmr} & 88.6 \\
EgoHMR~\cite{egohmr} & 85.8 \\
EgoPW~\cite{egopw} & 84.2 \\
Hong et al.~\cite{jswap} & 79.2 \\
EgoCast~\cite{egocast} & 78.1 \\
EgoTAP~\cite{egoattention} & 70.6 \\
EgoPoseFormer~\cite{egoposeformer} & 70.1 \\
Ours & \textbf{63.0} \\
\hline
\end{tabular}
\vspace{2mm}
\captionof{table}{Comparison on EgoPW~\cite{egopw} using PA-MPJPE (mm).}
\label{tab:egopw_comparison}
\end{minipage}
\hfill
\begin{minipage}[t]{0.48\textwidth}
\centering
\small
\begin{tabular}{l|c|c}
\hline
\textbf{Method} & \textbf{MPJPE}$\downarrow$ & \textbf{PA-MPJPE}$\downarrow$ \\
\hline
xR-EgoPose~\cite{xregopose} & 241.3 & 133.9 \\
EgoPW~\cite{egopw} & 189.6 & 105.3 \\
Hong et al.~\cite{jswap} & 131.9 & 94.6 \\
SceneEgo~\cite{sceneego} & 118.5 & 92.7 \\
EgoCast~\cite{egocast} & 123.5 & 82.4 \\
EgoPoseFormer~\cite{egoposeformer} & 102.2 & 77.3 \\
EgoTAP~\cite{egoattention} & 99.8 & 73.4 \\
Ours & \textbf{90.8} & \textbf{68.8} \\
\hline
\end{tabular}
\vspace{2mm}
\captionof{table}{Comparison on SceneEgo~\cite{sceneego} using MPJPE and
PA-MPJPE (mm).}
\label{tab:sceneego_comparison}
\end{minipage}

\end{table*}

\noindent\textbf{3D pose estimation training.}
We train on EgoPW and fine-tune the EgoPW checkpoint on SceneEgo. Unless stated
otherwise, clips have $T{=}64$ frames and stride 32. Frames are resized to
$256{\times}256$ for heatmaps and to $252{\times}252$ for DINOv2, yielding an
$18{\times}18$ ViT-S/14 patch grid. DINOv2 remains frozen. Heatmap-guided DINO
pooling uses temperature 6 and a 0.15 uniform fallback mixture; the
heatmap-statistics gate is constrained to $[0.05,0.80]$. For the
Ego4D-pretrained ActionFormer, we update only the input projection, embedding
convolutions, fusion projection, and last two branch blocks with a lower
learning rate than the newly initialized pose modules. We train for 32 epochs
with batch size 8 using AdamW (weight decay $10^{-4}$), an initial learning rate
of $2\times10^{-4}$ for the pose modules and $3\times10^{-5}$ for the adapted
ActionFormer parameters, three warm-up epochs followed by cosine decay to
$10^{-6}$, and gradient clipping at $\|\nabla\|_2\leq5.0$. The main loss uses
$\lambda_{\mathrm{sp}}{=}0.20$, $\lambda_{\mathrm{bone}}{=}0.10$,
$\lambda_{\mathrm{dir}}{=}0.10$, $\lambda_{\mathrm{res}}{=}0.005$, and
$\lambda_{\alpha}{=}0.001$.
\subsection{Evaluation Metrics}

We evaluate with standard 3D pose metrics. Mean Per Joint Position Error
(\textbf{MPJPE}) measures the average Euclidean distance between predicted and
ground-truth joints. \textbf{PA-MPJPE} first applies Procrustes alignment
(rotation, translation, and scale) and then computes MPJPE, measuring body
configuration independently of global pose and scale.

% Experiments
\section{Results}
\label{subsec:results}

We compare AG-EgoPose with prior egocentric 3D pose methods on EgoPW~\cite{egopw}
and SceneEgo~\cite{sceneego}, then ablate the main components of the current
residual-correction architecture.
\subsection{Quantitative Results}
For a fair comparison, recent stronger baselines with official implementations
(EgoPoseFormer~\cite{egoposeformer}, EgoTAP~\cite{egoattention}, and
EgoCast~\cite{egocast}) are retrained on EgoPW~\cite{egopw} and fine-tuned on
SceneEgo~\cite{sceneego} under the same protocol as our method. For
EgoPoseFormer, we use the monocular setting provided by the official code, so
no stereo-to-monocular architectural adaptation is introduced. Since Hong
et al.~\cite{jswap} do not provide official code, we include a paper-based
reimplementation following the architectural and training details reported in
their paper.

EgoPW labels are pseudo-ground truth recovered from synchronized external views
rather than marker-based capture, so their global position and scale do not
support an absolute metric; we therefore report only PA-MPJPE there and reserve
MPJPE for SceneEgo, which provides multi-view motion-capture annotations.
AG-EgoPose reaches \textbf{63.0}~mm PA-MPJPE on EgoPW
(Table~\ref{tab:egopw_comparison}), improving over
EgoPoseFormer~\cite{egoposeformer} by 7.1~mm and EgoTAP~\cite{egoattention} by
7.6~mm, with larger margins over EgoCast~\cite{egocast} and Hong
et al.~\cite{jswap}.

Fine-tuning the EgoPW-pretrained model on the SceneEgo training split and
evaluating on its held-out test split gives \textbf{90.8}~mm MPJPE and
\textbf{68.8}~mm PA-MPJPE (Table~\ref{tab:sceneego_comparison}), improving over
EgoTAP by 9.0 and 4.6~mm. Unlike SceneEgo, which relies on auxiliary depth and
semantic predictions, our model uses monocular RGB only.

\begin{figure*}[t]
\centering

\begin{tikzpicture}
\node[inner sep=0] (img)
 {\includegraphics[width=0.92\textwidth]{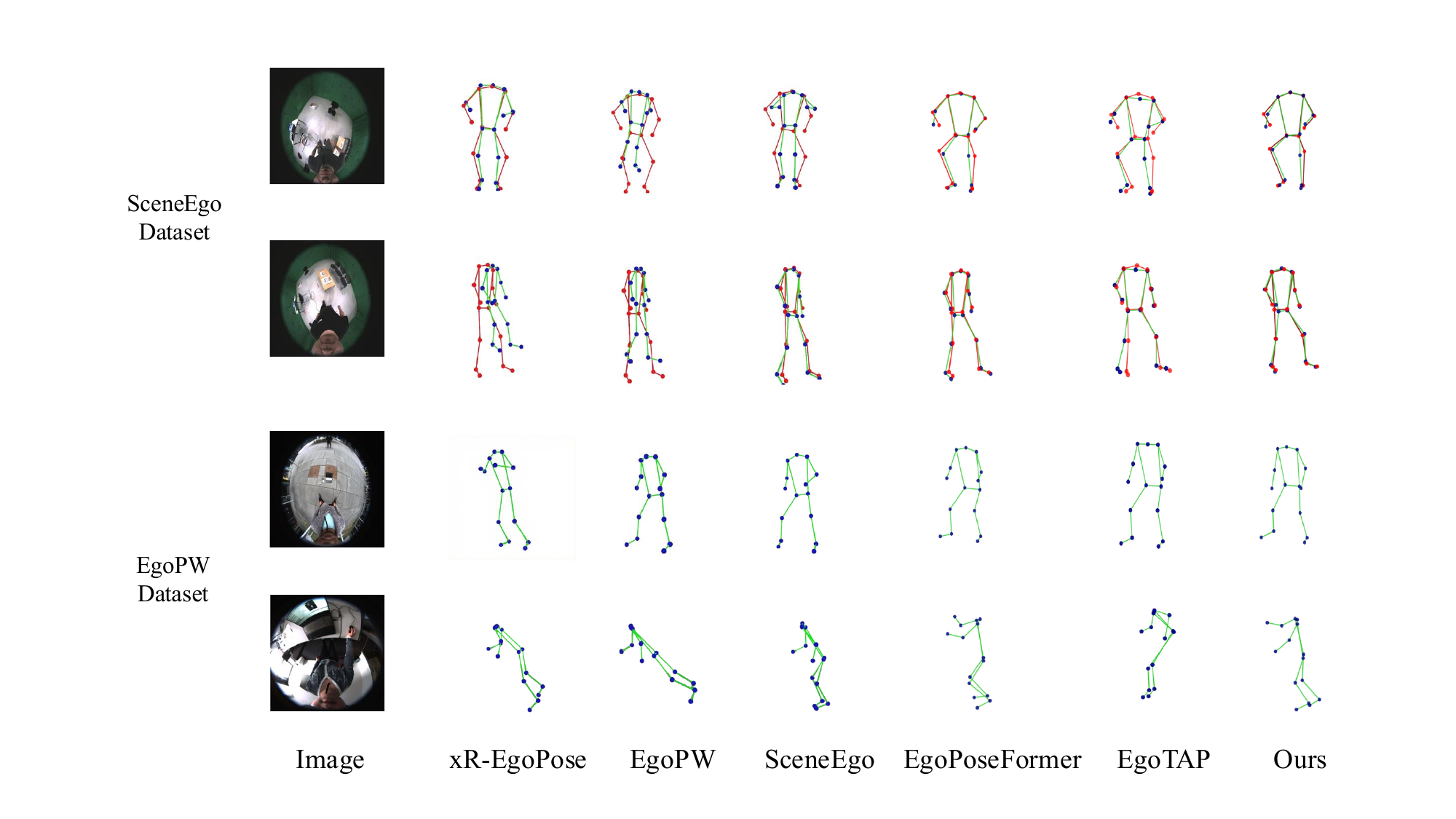}};

% --- Labels placed at fractional positions along the image width ---
% Tweak the 0.xx values until perfectly aligned with your columns.
% \node[anchor=north] at ($(img.south west)!0.172!(img.south east)$) {\footnotesize Image};
% \node[anchor=north] at ($(img.south west)!0.32!(img.south east)$) {\footnotesize Mo$^{2}$Cap$^{2}$};
% \node[anchor=north] at ($(img.south west)!0.46!(img.south east)$) {\footnotesize xR-EgoPose};
% \node[anchor=north] at ($(img.south west)!0.60!(img.south east)$) {\footnotesize EgoPW};
% \node[anchor=north] at ($(img.south west)!0.73!(img.south east)$) {\footnotesize SceneEgo};
% \node[anchor=north] at ($(img.south west)!0.85!(img.south east)$) {\footnotesize Ours};
\end{tikzpicture}
\caption{Qualitative comparison on SceneEgo~\cite{sceneego} and
EgoPW~\cite{egopw}. SceneEgo ground-truth poses are shown in red.}
\label{fig:qualitative}
\end{figure*}

\subsection{Qualitative Results}
Figure~\ref{fig:qualitative} shows controlled SceneEgo sequences and in-the-wild
EgoPW videos. Under fisheye distortion, self-occlusion, and partial visibility,
AG-EgoPose better preserves upper-body structure and leg articulation than prior
methods where direct image evidence is incomplete.

\begin{figure}[t]
\centering
\includegraphics[width=\linewidth]{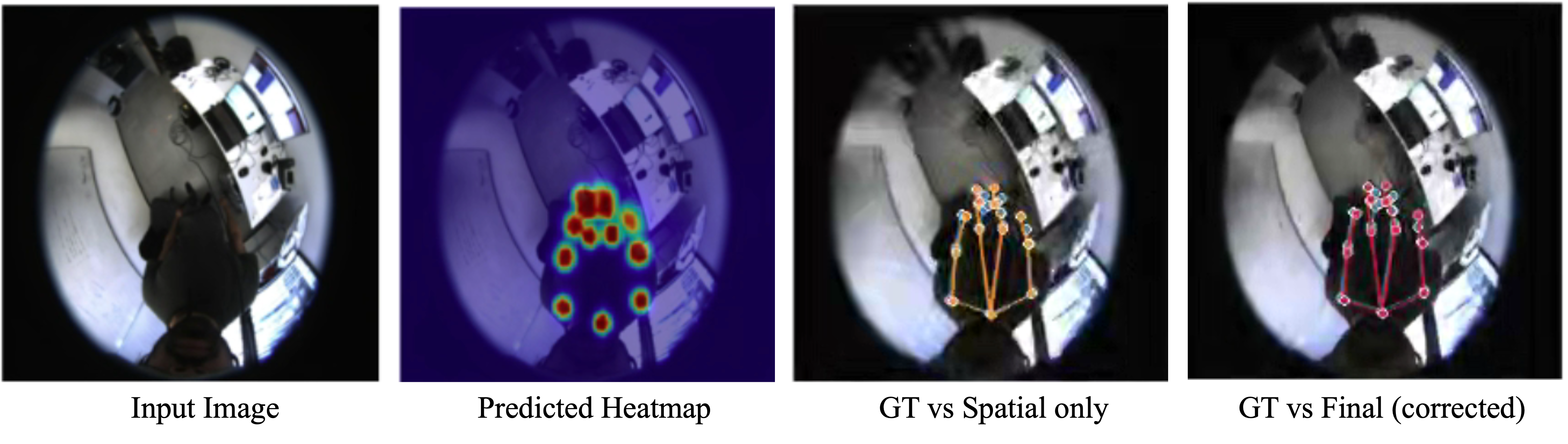}
\caption{Per-frame prediction flow: input, heatmap evidence, spatial anchor, and
temporally corrected pose. The correction reduces left/right wrist 3D error from
87.0~mm to 71.3~mm.}
\label{fig:qualitative_perframe}
\end{figure}

Figure~\ref{fig:qualitative_perframe} traces one difficult frame: the spatial
anchor remains uncertain around heavily occluded wrists, and the gated residual
draws on neighboring-frame context to refine them without replacing it.

\subsection{Ablation Studies}
\label{subsec:ablation}
We ablate the residual formulation using the same training and
fine-tuning protocol as the main model. EgoPW reports PA-MPJPE. SceneEgo
reports MPJPE and PA-MPJPE after fine-tuning from the corresponding EgoPW
checkpoint.

\begin{table}[t]
\centering
\footnotesize
\setlength{\tabcolsep}{4pt}
\renewcommand{\arraystretch}{1.1}
\begin{tabular}{lcc}
\toprule
\textbf{Variant} &
\begin{tabular}{@{}c@{}}\textbf{EgoPW}\\\textbf{PA-MPJPE}\end{tabular} &
\begin{tabular}{@{}c@{}}\textbf{SceneEgo}\\\textbf{MPJPE / PA-MPJPE}\end{tabular} \\
\midrule
Spatial anchor & 66.2 & 92.9 / 72.4 \\
Residual, $\alpha{=}1$ & 64.3 & 94.0 / 71.5 \\
Residual, fixed $\alpha{=}0.8$ & 63.7 & 92.9 / 71.8 \\
Temporal-feature gate & 63.8 & 92.5 / 71.4 \\
Heatmap-stat gate & \textbf{63.0} & \textbf{90.8 / 68.8} \\
\bottomrule
\end{tabular}
\caption{Effect of residual correction design. All residual variants refine the
same spatial anchor; the final model uses the heatmap-statistics gate.}
\label{tab:residual_anchor_ablation}
\end{table}

\begin{table}[t]
\centering
\small
\setlength{\tabcolsep}{4pt}
\renewcommand{\arraystretch}{1.1}
\begin{tabular}{lcc}
\toprule
\textbf{Variant} &
\begin{tabular}{@{}c@{}}\textbf{EgoPW}\\\textbf{PA-MPJPE}\end{tabular} &
\begin{tabular}{@{}c@{}}\textbf{SceneEgo}\\\textbf{MPJPE / PA-MPJPE}\end{tabular} \\
\midrule
Zero motion & 65.0 & 92.0 / 70.4 \\
Shuffled motion & 63.4 & 92.1 / 70.9 \\
Real motion & \textbf{63.0} & \textbf{90.8 / 68.8} \\
\bottomrule
\end{tabular}
\caption{Correction-signal ablation. Zero motion replaces the ActionFormer
output with zeros, testing residual decoder capacity; shuffled motion applies a
random permutation of that output along time within each clip, testing temporal
alignment. All rows use the same heatmap-statistics gate.}
\label{tab:correction_signal_ablation}
\end{table}

\noindent\textbf{Residual correction.}
Every residual variant improves EgoPW over the spatial anchor
(Table~\ref{tab:residual_anchor_ablation}), but SceneEgo separates them. An
ungated residual is actively harmful there, costing 1.1~mm of MPJPE relative to
the anchor, and a fixed gate merely recovers it. Learned gates do better, and
conditioning on heatmap statistics rather than on temporal features is worth a
further 1.7~mm MPJPE and 2.6~mm PA-MPJPE. Correction strength therefore has to
be modulated, and modulating it from spatial evidence transfers best.

\noindent\textbf{Correction signal.}
Table~\ref{tab:correction_signal_ablation} asks what the residual actually needs.
Removing motion entirely leaves only 1.2~mm of the 3.2~mm EgoPW gain and 0.9~mm
of the 2.1~mm on SceneEgo MPJPE, so the decoder and its joint-local appearance
features recover less than half of the correction on their own; real
ActionFormer context supplies the larger remainder, 2.0~mm on EgoPW and 1.2~mm
on SceneEgo. Shuffling those features along time separates their content from
their ordering. In-domain, scrambled context still recovers 1.6 of those 2.0~mm,
so part of the benefit is representational rather than temporal. Under transfer
the ordering dominates: scrambled context is worse than no context at all, and
restoring the correct order is worth 1.3~mm MPJPE and 2.1~mm PA-MPJPE.

\begin{table}[t]
\centering
\footnotesize
\setlength{\tabcolsep}{3pt}
\renewcommand{\arraystretch}{1.1}
\begin{tabular}{lcc}
\toprule
\textbf{Temporal encoder} &
\begin{tabular}{@{}c@{}}\textbf{EgoPW}\\\textbf{PA-MPJPE}\end{tabular} &
\begin{tabular}{@{}c@{}}\textbf{SceneEgo}\\\textbf{MPJPE / PA-MPJPE}\end{tabular} \\
\midrule
Matched TCN & 67.3 & 99.5 / 73.8 \\
Random ActionFormer & 64.8 & 93.2 / 70.1 \\
Frozen ActionFormer & 64.5 & 91.9 / 69.4 \\
Fine-tuned ActionFormer & \textbf{63.0} & \textbf{90.8 / 68.8} \\
\bottomrule
\end{tabular}
\caption{Temporal encoder ablation. The TCN controls for temporal capacity
without action pretraining, while the ActionFormer variants separate architecture,
pretraining, and task adaptation.}
\label{tab:temporal_encoder_ablation}
\end{table}

\noindent\textbf{Temporal encoder and action prior.}
Table~\ref{tab:temporal_encoder_ablation} separates three factors that are often
conflated: temporal capacity, architecture, and the transferred prior. The
matched TCN is worst by a wide margin, so generic temporal smoothing does not
supply what the residual branch needs. Replacing it with a randomly initialized
ActionFormer recovers 2.5~mm on EgoPW, which is attributable to the multi-scale
attention architecture alone. Ego4D pretraining adds a further gain that is
concentrated on SceneEgo, the smaller and more distribution-shifted benchmark,
and adapting the last two blocks is better still. The prior therefore matters
most where target data is scarce.

\begin{table}[t]
\centering
\small
\setlength{\tabcolsep}{4pt}
\renewcommand{\arraystretch}{1.1}
\begin{tabular}{lcc}
\toprule
\textbf{Variant} &
\begin{tabular}{@{}c@{}}\textbf{EgoPW}\\\textbf{PA-MPJPE}\end{tabular} &
\begin{tabular}{@{}c@{}}\textbf{SceneEgo}\\\textbf{MPJPE / PA-MPJPE}\end{tabular} \\
\midrule
Full model & \textbf{63.0} & \textbf{90.8 / 68.8} \\
No FPN token & 64.2 & 94.3 / 74.0 \\
No SJT & 68.9 & 104.8 / 77.7 \\
No gate stats & 63.5 & 92.0 / 71.1 \\
\bottomrule
\end{tabular}
\caption{Spatial evidence ablation. ``No FPN token'' drops the sampled FPN
feature from the tokenizer; ``No gate stats'' zeroes $\mathbf{u}_{t,j}$ at the
gate MLP, leaving a single learned correction weight.}
\label{tab:spatial_evidence_ablation}
\end{table}

\noindent\textbf{Heatmap-grounded spatial evidence.}
Table~\ref{tab:spatial_evidence_ablation} isolates the spatial signals. Dropping
the Spatial Joint Transformer is by far the most damaging change, costing
14.0~mm of SceneEgo MPJPE: without cross-joint reasoning the tokenizer output
remains a set of independent per-joint descriptors, and left/right and kinematic
errors survive into the anchor. The FPN sample and the gate statistics cost less
individually, but both losses fall almost entirely on SceneEgo, where the
transfer gap is largest.

\begin{table}[t]
\centering
\footnotesize
\setlength{\tabcolsep}{3pt}
\renewcommand{\arraystretch}{1.1}
\begin{tabular}{lcc}
\toprule
\textbf{Variant} &
\begin{tabular}{@{}c@{}}\textbf{EgoPW}\\\textbf{PA-MPJPE}\end{tabular} &
\begin{tabular}{@{}c@{}}\textbf{SceneEgo}\\\textbf{MPJPE / PA-MPJPE}\end{tabular} \\
\midrule
MPJPE only & 65.2 & 97.1 / 74.2 \\
MPJPE + spatial & 64.6 & 97.4 / 74.1 \\
MPJPE + bone/dir. & 64.5 & 94.9 / 73.0 \\
MPJPE + spatial + bone/dir. & 63.8 & 92.8 / 72.3 \\
Full objective & \textbf{63.0} & \textbf{90.8 / 68.8} \\
\bottomrule
\end{tabular}
\caption{Training objective ablation. This study evaluates the roles of spatial
anchor supervision, kinematic losses, and correction regularization.}
\label{tab:loss_ablation}
\end{table}

\noindent\textbf{Training objective.}
Spatial-anchor supervision mainly helps EgoPW, while the bone-length and
limb-direction terms help both datasets by enforcing plausible limb geometry
(Table~\ref{tab:loss_ablation}). The two are additive, and the small residual
and $\alpha$ penalties contribute the last 0.8~mm on EgoPW and 3.5~mm on SceneEgo
PA-MPJPE, most of it under transfer.

\begin{table}[t]
\centering
\footnotesize
\setlength{\tabcolsep}{3pt}
\renewcommand{\arraystretch}{1.1}
\begin{tabular}{lcc}
\toprule
\textbf{DINO pooling} &
\begin{tabular}{@{}c@{}}\textbf{EgoPW}\\\textbf{PA-MPJPE}\end{tabular} &
\begin{tabular}{@{}c@{}}\textbf{SceneEgo}\\\textbf{MPJPE / PA-MPJPE}\end{tabular} \\
\midrule
CLS only & 64.1 & 94.2 / 73.1 \\
Mean patch & 63.8 & 93.5 / 71.8 \\
Heatmap patch & 63.2 & 91.5 / 70.6 \\
\quad + uniform prior & \textbf{63.0} & \textbf{90.8 / 68.8} \\
\bottomrule
\end{tabular}
\caption{DINO patch pooling ablation, testing whether the residual decoder needs
joint-local appearance or only a global descriptor. Rows follow
Eq.~\eqref{eq:dino_pooling} with $\rho{=}0$ and $\rho{=}0.15$.}
\label{tab:dino_pooling_ablation}
\end{table}

\noindent\textbf{Joint-local DINO pooling.}
Global visual features are not sufficient for the residual branch
(Table~\ref{tab:dino_pooling_ablation}). CLS-only pooling is weakest, and mean
patch pooling recovers little, since neither gives a joint its own descriptor.
Heatmap-guided pooling does, and the uniform prior adds a further 0.7~mm of
SceneEgo MPJPE by preventing a displaced heatmap from committing to one wrong
patch.

\begin{table}[t]
\centering
\small
\setlength{\tabcolsep}{4pt}
\renewcommand{\arraystretch}{1.1}
\begin{tabular}{lcc}
\toprule
\textbf{Sequence length} &
\begin{tabular}{@{}c@{}}\textbf{EgoPW}\\\textbf{PA-MPJPE}\end{tabular} &
\begin{tabular}{@{}c@{}}\textbf{SceneEgo}\\\textbf{MPJPE / PA-MPJPE}\end{tabular} \\
\midrule
16 & 65.5 & 97.2 / 73.2 \\
32 & 64.5 & 94.0 / 71.3 \\
64 & \textbf{63.0} & 90.8 / \textbf{68.8} \\
128 & 63.1 & \textbf{89.7} / 69.9 \\
\bottomrule
\end{tabular}
\caption{Sequence-length ablation. Clip length in frames, all other settings
fixed.}
\label{tab:sequence_length_ablation}
\end{table}

\noindent\textbf{Temporal context length.}
Table \ref{tab:sequence_length_ablation} shows short clips are clearly worse, so the residual branch needs more than
frame-to-frame evidence. Doubling to 128 frames trades 1.1~mm of SceneEgo MPJPE
for 1.1~mm of PA-MPJPE and does not help EgoPW, so we keep 64 frames.

% \begin{table}[t]
% \centering
% \small
% \setlength{\tabcolsep}{5pt}
% \renewcommand{\arraystretch}{1.1}
% \begin{tabular}{lccc}
% \toprule
% \textbf{Joint} & \textbf{Spatial anchor} & \textbf{Full model} & $\bar{\alpha}$ \\
% \midrule
% Wrist    & 106.5 & \textbf{94.7} & 0.425 \\
% Hip      &  62.0 & \textbf{58.8} & 0.413 \\
% Knee     &  67.2 & \textbf{64.8} & 0.420 \\
% Foot     &  75.7 & \textbf{74.1} & 0.423 \\
% Ankle    &  67.3 & \textbf{65.8} & 0.421 \\
% Shoulder &  41.2 & \textbf{39.8} & 0.410 \\
% Neck     &  41.4 & \textbf{40.1} & 0.399 \\
% Elbow    &  58.5 & \textbf{57.3} & 0.420 \\
% \bottomrule
% \end{tabular}
% \caption{Per-joint EgoPW error (mm) before and after residual correction, with
% the mean gate value. Left and right joints are averaged.}
% \label{tab:per_joint}
% \end{table}

% \noindent\textbf{Per-joint behavior.}
% Table~\ref{tab:per_joint} shows that the correction is not spread evenly. Wrists
% are by far the hardest joints for the spatial anchor and receive by far the
% largest correction, 11.8~mm, while well-localized joints such as the neck and
% elbow move by roughly 1~mm. The mean gate follows the same ordering: across
% joints, $\bar{\alpha}$ correlates with spatial-anchor error at $r{=}0.79$,
% rising for wrists and falling for the neck. The gate therefore allows more
% correction exactly where frame-level evidence is weakest, which is the behavior
% the formulation was designed to produce.

% Conclusion
\section{Conclusion}
\label{sec:conclusion}

We propose AG-EgoPose, a residual formulation for egocentric 3D human pose
estimation that separates reliable spatial pose estimation from bounded-gate
residual correction. The model builds a supervised pose anchor from
heatmap-grounded spatial evidence and uses ActionFormer features as a
complementary action-context signal. This design reaches 63.0 mm PA-MPJPE on
EgoPW and 90.8/68.8 mm MPJPE/PA-MPJPE on SceneEgo. The spatial anchor accounts
for most of the raw accuracy, with the residual adding roughly 3 mm on top of
it. Within that correction, the temporal branch supplies the larger share:
removing motion recovers less than half of it on either benchmark. Part of this
benefit survives temporal shuffling in-domain, so
it is not purely a matter of frame ordering, but under transfer the correct
ordering accounts for most of it.

Our method also has limitations. The current model is evaluated in an offline
sliding-window setting and is not strictly causal; a real-time deployment would
require a causal temporal encoder or an explicit latency budget. The
heatmap-statistics gate controls correction from the spatial state, but it does
not directly measure temporal reliability. Future work could incorporate
temporal consistency, visibility cues, and causal action context, especially for
settings with different actions, camera geometry, or body visibility.

{\small
\bibliographystyle{ieeenat_fullname}
\bibliography{main}
}
% Supplementary material lives in its own driver, main_suppl.tex, so that it
% compiles to a separate PDF and the submitted main paper stays unchanged.
% \input{sec/X_suppl}
\clearpage
\setcounter{page}{1}
\maketitlesupplementary

% Supplementary sections become A, B, ...; figures and tables become S1, S2, ...
% so that references are never ambiguous with the main paper.
\setcounter{section}{0}
\renewcommand{\thesection}{\Alph{section}}
\setcounter{figure}{0}
\renewcommand{\thefigure}{S\arabic{figure}}
\setcounter{table}{0}
\renewcommand{\thetable}{S\arabic{table}}

\noindent
The main paper evaluates the correction gate only through aggregate metrics.
Here we examine what the gate learns (\cref{sec:supp_gate}) and how the residual
correction is distributed across joints (\cref{sec:supp_perjoint}), extending
Sec.~3.3.2 and Sec.~5.3 of the main paper. All numbers are obtained by inference
on the held-out test splits using the checkpoints reported in the main paper; no
additional training was performed.

\section{Gate Behavior}
\label{sec:supp_gate}

\begin{figure*}[t]
\centering
\includegraphics[width=0.92\textwidth]{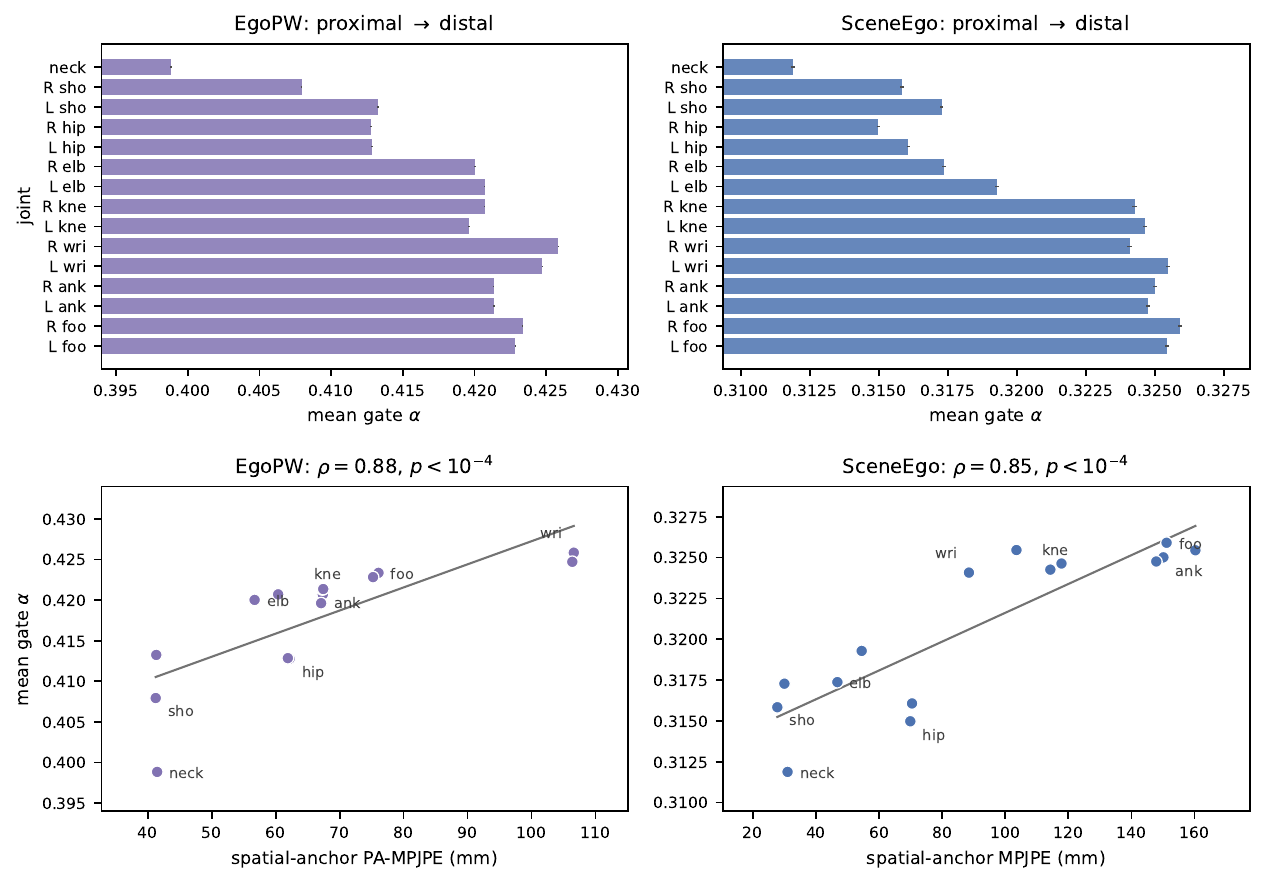}
\caption{Per-joint gate behavior. Each bar or point is one joint, using the mean
gate value and mean spatial-anchor error over the test split. EgoPW reports
PA-MPJPE and SceneEgo MPJPE, following the metric convention of the main paper.
Top: joints ordered proximal to distal. Bottom: the same gate values against
measured anchor error, with Spearman correlations. Error bars are standard
errors of the mean and are smaller than the bar edges.}
\label{fig:supp_gate}
\end{figure*}

Eq.~(6) of the main paper conditions the correction weight on heatmap
statistics, with the intent of spending correction where frame-level evidence is
weak. Figure~\ref{fig:supp_gate} asks whether the trained gate does this. For
each joint we record the mean gate value $\bar{\alpha}_j$ and the mean error of
the spatial anchor $\mathbf{P}^{\mathrm{sp}}$ over the full test split
($99{,}072$ frames on EgoPW, $13{,}120$ on SceneEgo).

\paragraph{Gate ordering.}
The ordering matches the intent. Joints that the heatmap network localizes
reliably receive the smallest correction weight, and the least reliable joints
receive the largest: on EgoPW the gate is lowest at the neck ($0.399$) and
highest at the wrists ($0.425$), which have the smallest and largest anchor
error respectively. The rank correlation between $\bar{\alpha}_j$ and anchor
error is $\rho{=}0.88$ on EgoPW and $\rho{=}0.85$ on SceneEgo ($p<10^{-4}$), and
the two benchmarks agree despite differing in which joints are hardest, so the
effect is not specific to one test set.

\paragraph{Gate magnitude.}
The gate occupies only a small part of the interval it is allowed. It averages
$0.418$ ($\sigma{=}0.019$) on
EgoPW and $0.321$ ($\sigma{=}0.010$) on SceneEgo, spanning $0.399$--$0.426$ and
$0.312$--$0.326$ across joints, and neither bound of the permitted interval
$[0.05,0.80]$ is reached at any joint or frame. The bounds in Eq.~(6) therefore
constrain optimization rather than the converged model. In practice the gate acts
as a bounded scale on the correction rather than a per-sample switch: it sets how
much correction is applied, while \cref{sec:supp_perjoint} shows that where it is
applied is determined by $\Delta\mathbf{P}$.

\section{Per-Joint Correction}
\label{sec:supp_perjoint}

Table~\ref{tab:supp_perjoint} reports the spatial anchor and the full model per
joint on EgoPW, together with the mean gate. Left and right joints are averaged.

\begin{table}[t]
\centering
\small
\setlength{\tabcolsep}{5pt}
\renewcommand{\arraystretch}{1.1}
\begin{tabular}{lccc}
\toprule
\textbf{Joint} & \textbf{Spatial anchor} & \textbf{Full model} & $\bar{\alpha}$ \\
\midrule
Wrist    & 106.5 & \textbf{94.7} & 0.425 \\
Foot     &  75.7 & \textbf{74.1} & 0.423 \\
Ankle    &  67.3 & \textbf{65.8} & 0.421 \\
Knee     &  67.2 & \textbf{64.8} & 0.420 \\
Hip      &  62.0 & \textbf{58.8} & 0.413 \\
Elbow    &  58.5 & \textbf{57.3} & 0.420 \\
Neck     &  41.4 & \textbf{40.1} & 0.399 \\
Shoulder &  41.2 & \textbf{39.8} & 0.410 \\
\bottomrule
\end{tabular}
\caption{Per-joint EgoPW PA-MPJPE (mm) before and after residual correction,
with the mean gate value. Rows are ordered by anchor error; left and right
joints are averaged.}
\label{tab:supp_perjoint}
\end{table}

\paragraph{Distribution across joints.}
The correction is distributed unevenly. Wrists are by far the hardest joints for
the spatial anchor at $106.5$~mm and receive the largest correction, $11.8$~mm,
whereas joints the anchor already localizes well move little: $1.3$~mm at the
neck and $1.2$~mm at the elbow. This matches the qualitative example in the main
paper, where the correction acts mainly on occluded wrists.

\paragraph{Gate versus residual.}
The concentration is not produced by the gate. Since
$\mathbf{P}=\mathbf{P}^{\mathrm{sp}}+\alpha\Delta\mathbf{P}$, the correction
applied to a joint is the product of its gate and its residual. Wrists receive
$9.1\times$ more correction than the neck, but their gate is only $1.07\times$
larger; the remaining factor of $8.5$ comes from the magnitude of
$\Delta\mathbf{P}$. The residual decoder thus determines which joints to correct
and in which direction, while the gate contributes a bounded scale tied to
spatial rather than temporal evidence. This is consistent with the ablation in
the main paper, where removing the gate statistics costs $2.3$~mm of SceneEgo
PA-MPJPE but only $0.5$~mm on EgoPW.

\end{document}